\pdfoutput=1
\documentclass[11pt]{article}

\usepackage[final]{acl}

\usepackage{times}
\usepackage{latexsym}
\usepackage{graphicx}
\usepackage{amsmath}
\usepackage{amssymb}
\usepackage{color}

\usepackage[T1]{fontenc}

\usepackage[utf8]{inputenc}

\usepackage{microtype}
\usepackage{enumitem}
\DeclareMathOperator*{\argmax}{argmax}

\newcommand*{\affaddr}[1]{#1} 
\newcommand*{\affmark}[1][*]{\textsuperscript{#1}}
\newcommand*{\email}[1]{\text{#1}}

\title{Learn To Remember: Transformer with Recurrent Memory

for Document-Level Machine Translation}

\author{%
Yukun Feng\affmark[1], Feng Li\affmark[2], Ziang Song\affmark[1], Boyuan Zheng\affmark[1], and Philipp Koehn\affmark[1]\\
\affaddr{\affmark[1]Department of Computer Science, Johns Hopkins University}\\
\affaddr{\affmark[2]Department of Computer Science, University of Illinois Urbana-Champaign}\\
\email{\{yfeng55, zsong17, bzheng12, phi\}@jhu.edu, \{fengl3\}@illinois.edu}
}

\begin{document}
\maketitle

\begin{abstract}
The Transformer architecture has led to significant gains in machine translation. However, most studies focus on only sentence-level translation without considering the context dependency within documents, leading to the inadequacy of document-level coherence.
Some recent research tried to mitigate this issue by introducing an additional context encoder or translating with multiple sentences or even the entire document.
Such methods may lose the information on the target side or have an increasing computational complexity as documents get longer.
To address such problems, we introduce a recurrent memory unit to the vanilla Transformer, which supports the information exchange between the sentence and previous context. The memory unit is recurrently updated by acquiring information from sentences, and passing the aggregated knowledge back to subsequent sentence states.
We follow a two-stage training strategy, in which the model is first trained at the sentence level and then finetuned for document-level translation.
We conduct experiments on three popular datasets for document-level machine translation and our model has an average improvement of 0.91 s-BLEU over the sentence-level baseline. We also achieve state-of-the-art results on TED and News,  outperforming the previous work by 0.36 s-BLEU and 1.49 d-BLEU on average.
\end{abstract}

\section{Introduction}
Most previous machine translation methods are designed for sentence-level translation. 
Recent studies have shown that the effective use of contextual information between sentences can achieve better performance in document-level machine translation \cite{garcia-etal-2015-document-level, maruf-haffari-2018-document, 
miculicich-etal-2018-document,
zhang-etal-2020-long, bao-etal-2021-g}. 
Built on the Transformer model \cite{NIPS2017_3f5ee243}, a general approach is to incorporate neighboring sentence states  \cite{tiedemann-scherrer-2017-neural, ijcai2020-0551} into the attention mechanism, which has also been widely used in many long sequence modeling methods \cite{dai-etal-2019-transformer,Rae2020Compressive,Yang2019XLNetGA,Beltagy2020Longformer}. 
\citet{zhang-etal-2018-improving, maruf-etal-2019-selective} have introduced an additional context encoder to solve the limitation of sentence-level translation, which, however, is separated from the original translation model and context states is only applied on the source side.
Other works \cite{junczys-dowmunt-2019-microsoft, scherrer-etal-2019-analysing, zhang-etal-2020-long, bao-etal-2021-g} concatenated sentences or the entire document and feed into the attention module of the Transformer. Since more extended contexts may confound attention on meaningful portions of the current sentence, the model is difficult to select valuable inputs from extra contexts to navigate the redundancy of information. Such methods also suffer from the quadratically increasing complexity when documents get longer. 


We solve such problems by introducing a memory mechanism to recurrently integrate contextualized knowledge from intermediate state in Transformer layers. As sentences are ordered in documents, our model reads one sentence pair at each step, keeping the computational cost as same as the sentence-level translation.
As recurrent memory has been widely researched since RNN \cite{Rumelhart:1986we}, which has been incorporated with Transformer by Transformer-XL \cite{dai-etal-2019-transformer} and further extended by \citet{Rae2020Compressive} who compress previous states into a two-layer hidden memory.
In our approach, we update the memory through an attention module to select practical information from sentences and reduce the context space into multiple dense vectors in the memory.
Besides, we use another attention module to pass the knowledge retained in the memory back to the sentence state in the next step. 
Such information exchange is expected to convey contextualized dependency between sentences.
This memory mechanism can be applied in each layer for both the source and target documents, and our study shows that incorporating memory only in the last layer achieves the best performance. 

We experiment across three widely used datasets for document-level translation: TED, NEWS, and Europarl, and evaluate our model with s-BLEU and d-BLEU.
We first train a vanilla Transformer on sentence-level translation as the baseline and finetune the model for the documents by initializing the memory mechanism to the Transformer.
Our model outperforms previous SOTA work by 0.5 s-BLEU and 2.30 d-BLEU on TED, and 0.21 s-BLEU and 0.57 d-BLEU on News.
We do not achieve the SOTA result on Europarl, which might be caused by the different results between the baselines for sentence-level translation.
However, we further evaluate the improvement of previous works from their reported baseline Transformer, and we achieve the most relative gain on all three datasets.
We also analyze our model from the memory usage, long-range effect, context dependency, and computational complexity, and demonstrate the effectiveness and efficiency of our approach in the general understanding of the document machine translation.

Overall, this paper makes several contributions: 
\textbf{(i)} Our work reduces the contextualized knowledge space of sentences states to multiple dense vectors, and considers the sentence dependency for both source and target documents, while keeping computational complexity in sentence-level. 
\textbf{(ii)} Our model significantly improves the sentence level baseline by 0.91 s-BLEU average and achieved the SOTA results on TED and News.
\textbf{(iii)} Our model shows the effective use of memory, long-range influence, context-dependency across sentences, and decoding efficiency through convincing analysis.

\begin{figure*}[t]
\includegraphics[scale=0.13]{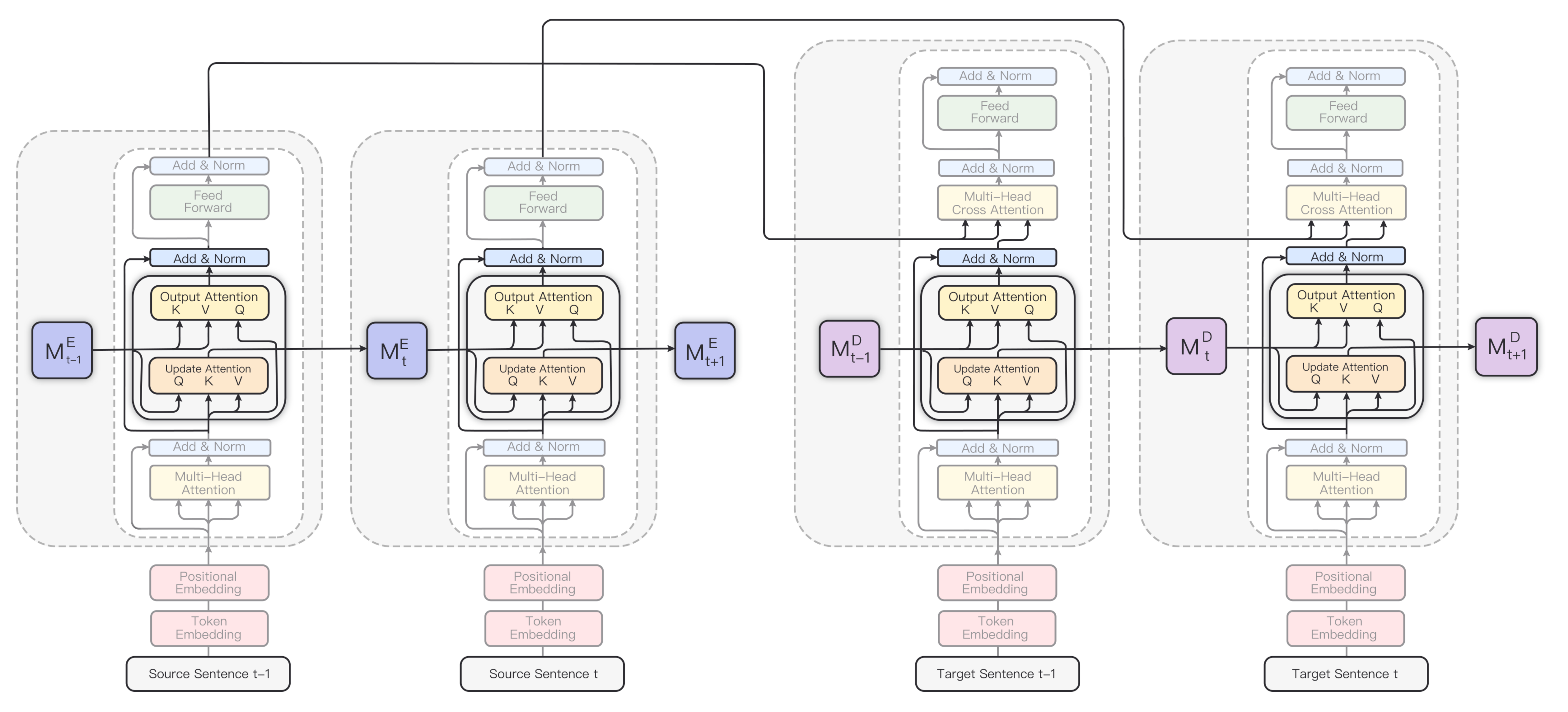}
\centering
\caption{An overview of the model architecture, where $\mathrm{E}$ and $\mathrm{D}$ refers to Encoder and Decoder respectively.}
\label{tram-model}
\end{figure*}


\section{Related Works}
\paragraph{Recurrent Sequence Modeling}
RNN \cite{Rumelhart:1986we} was the first class of models that introduced hidden states as the memory in neural models. Although improved on sequential-oriented tasks, RNN has unsatisfactory learning of long-term information due to gradient vanishing and explosion. LSTM \cite{hochreiter1997long} improved RNN by introducing gate mechanisms to selectively retain knowledge at each step. 

This RNN variant dominated NLP models until the Transformer replaced the memory unit with a self-attention mechanism and achieved great success in a wide range of NLP applications. Although we cannot deny the robustness and effectiveness of the Transformer model, the quadratically increased computational cost as the increase of token numbers makes Transformer unable to fit the long-range sequence.
Some studies \cite{imageTrans2018,child2019generating, Beltagy2020Longformer,ainslie-etal-2020-etc,qiu-etal-2020-blockwise,zaheer2020bigbird,martins2021inftyformer} try to mitigate this issue by reducing the complexity of the attention module. 
However, such work still suffers from the problems by unlimited the document length and the document modeling is hard to solve. 

Transformer-XL \cite{dai-etal-2019-transformer} broke this dilemma by introducing the recurrent memory into Transformer-based models. 
It cached previous hidden sentences computation and mapped such states to subsequent sentences states. Theoretically, Transformer-XL could handle infinite length text but storing uncompressed hidden state requires tremendous memory space, which impeded Transformer-XL from good performance on dealing with practical long-sequence tasks.  The Compressive Transformer \cite{Rae2020Compressive} further addressed this problem by mapping the evicted hidden state from cached memory to a more compressed representation.
However, two-layer caching still requires a huge memory space and may be improved with trainable memories.

\paragraph{Document Machine Translation}
Machine Translation has been a widely researched area for decades. A series of models have addressed various translation problems \cite{koehn-etal-2003-statistical, kalchbrenner-blunsom-2013-recurrent, Bahdanau2015NeuralMT, luong-etal-2015-effective}. 
As most of them target translation at the sentence level, document-level translation poses a fundamental challenge requiring models to pass intra-sentential information throughout consecutive sequences of sentences, and it has been addressed by \citet{gong-etal-2011-cache, hardmeier-etal-2013-docent, pouget-abadie-etal-2014-overcoming, garcia-etal-2015-document-level, koehn-knowles-2017-six, laubli-etal-2018-machine, Agrawal2018ContextualHI} among others. 

Recent studies have attempted to incorporate additional contextual information into the Transformer structure to improve the performance of neural machine translation models further. 
The intuitive way is to leverage neighboring sentences from paragraphs or the documents \cite{tiedemann-scherrer-2017-neural, maruf-haffari-2018-document, ijcai2020-0551}, demonstrating the effectiveness of the additional contexts. 
Specifically, in the first class of methodologies for document-level translation, independent from the architecture of vanilla Transformer processing current sentences, some studies \cite{miculicich-etal-2018-document, zhang-etal-2018-improving, maruf-etal-2019-selective, voita-etal-2019-context, voita-etal-2019-good, ma-etal-2020-simple, donato-etal-2021-diverse} introduces context-aware components only attend to source or target contexts and usually jointly train with the rest of the network from scratch. 
The second class of models follows the pattern of concatenating multiple sentences for translation
\cite{Agrawal2018ContextualHI, scherrer-etal-2019-analysing, junczys-dowmunt-2019-microsoft, zhang-etal-2020-long}. 
Such a method is expected to capture the contextual correlations between sentences.
However, one of its drawbacks is the quadratically increased computational complexity in the face of longer contexts sequences. 
Also, longer sequences usually confound document-level attention and sometimes even overlook key information on the current sentences. 
\citet{bao-etal-2021-g} uses group masks to introduce locality constraints to reinforce sentence information in multi-head attention to resolve the confounding issue in long contexts.


Our work incorporates the idea of the recurrent memory to document-level machine translation. It follows the locality assumptions by reducing the context space into multiple memory vectors and passes dependencies between sentences. 
The mechanism to update and output memory is similar to models  which store cached bilingual sentence pairs in the memory to enhance the sentence-level translation \cite{feng-etal-2017-memory, he-etal-2021-fast, jiang2021documentlevel}. We believe our approach is intuitive to efficiently store sentence states and transfer context information across sentences. 



\section{Approach}

Our model is shown in Figure \ref{tram-model}.
Additional to the vanilla Transformer, we introduce a contextual memory unit and two attention modules to manipulate the memory defined as Update Attention and Output Attention. These modules can be applied at each layer in both the encoder and decoder.

As input sentences are ordered from left to right in the document, our model only reads one sentence every time.
The memory is expected to store contextualized information from the input sentence states and convey such knowledge to the next sentence.
At each step, the Update Attention step maps the contextual information from the sentence state to the memory, and updates the memory to the next step.
Meanwhile, the Output Attention step fuses the information from the current sentence and the contextual memory, and outputs the aggregated knowledge to the remaining modules of the layer.

Formally, we define $\mathrm{h^t}$ as the sentence state from self-attention module in Transformer layer, and $\mathrm{M^t}$ refers to the contextual memory $\mathrm{M}$ at step $\mathrm{t}$, where $\mathrm{t}$ refers to the index of $\mathrm{t^{th}}$ sentence in the document.
$\mathrm{M^t}$ and $\mathrm{h^t}$ are updated and outputted as:
\begin{equation}
\begin{split}
\mathrm{M^{t+1} = UpdateAttention(M^t, h^t)} \\
\mathrm{\widetilde{h^t} = OutputAttention(M^t, h^t)}
\end{split}
\end{equation} 




\subsection{Contextual Memory}
Memory $\mathrm{M} \in $
$\mathbb{R}^\mathrm{{d_{M} \times d_{model}}}$
where $\mathrm{d_{model}}$ refers to the hidden dimension and $\mathrm{d_{M}}$ is a hyper-parameter, indicating how many vectors will be allocated for memory. To avoid the redundancy of memory space, we set $\mathrm{d_{M}}$ to 16. Detailed analysis is discussed later.

\subsection{Update Attention}
We update contextual memory through an attention module \cite{NIPS2017_3f5ee243}. Attention is a mapping function between input vectors of query ($\mathrm{Q}$) and key-value ($\mathrm{K}$-$\mathrm{V}$) pairs. 
The output is the weighted sum of values with corresponding scores.
\begin{equation*}
\mathrm{Attention(Q,K,V) = Softmax(\frac{QK^T}{\sqrt{d_k}})V}
\end{equation*}
Multi-Head attention extends the vanilla attention by projecting input vectors ($\mathrm{Q,K,V}$) to different representation subspaces, and attention is performed in parallel in each head. Attention outputs from multiple heads will be concatenated and projected to the expected space.
\begin{equation*}
\begin{split}
\mathrm{MHA(Q,K,V) = Concat(head_1,..,head_n)W^o} \\
\mathrm{head_i = Attention(QW_i^q,KW_i^k,VW_i^v)}
\end{split}
\end{equation*}
where $\mathrm{d_k}$ is the hidden dimension of the $\mathrm{K}$,
$\mathrm{W^q}$, $\mathrm{W^k}$, $\mathrm{W^v}$ $\in$
$\mathbb{R}^\mathrm{d_{model} \times {d_h}}$, 
and $\mathrm{W^o}$ $\in$ $\mathbb{R}^\mathrm{n\times{d_{h}}\times{d_{model}}}$ are learnable parameters. $\mathrm{d_{model}}$ and $\mathrm{d_h}$ refer to the hidden dimension of the model and each head.

To update the contextual memory $\mathrm{M^t}$ to next step, sentence state $\mathrm{h^t}$ is mapped to $\mathrm{M^t}$ through the Multi-Head Attention. Both the memory and context state are projected into different sub-spaces and contextualized knowledge is expected to be mapped to each memory vector from different perspectives. The memory at step $\mathrm{t}$ is updated as:
\begin{equation}
\mathrm{\widetilde{M^t} = AddNorm(MHA(M^t, h^t, h^t))}
\end{equation} 

A Feed-Forward Network is then used to further enhance the memory representation from the attention output. 
\begin{equation}
\mathrm{M^{t+1} = AddNorm(FeedForward( \widetilde{M^t}))}
\end{equation}

In the memory matrix $\mathrm{M}$, each vector is expected to select contextualized information from different perspectives. However, it is hard to distinguish such vectors since they do not have actual positional meanings, and the same key-value pairs are mapped to these vectors in the attention phase resulting in the same representation in each memory vector.
To solve such a problem, we use the
positional encoding  $\mathrm{PE()}$ as introduced in \citet{NIPS2017_3f5ee243} to differentiate multiple memory vectors. $\mathrm{M}$ is added by such position-level bias in each update phase. 
\begin{equation}
\mathrm{M^{t} = M^{t} + PE(M^{t})}
\end{equation}

\subsection{Output Attention}
To map the contextualized knowledge from $\mathrm{M^t}$ to the sentence state $\mathrm{h^t}$, multi-head attention is used to take the representation of $\mathrm{h^t}$ and $\mathrm{M^t}$ as query and key-value, respectively. 
\begin{equation}
\mathrm{\widetilde{h^t} = MHA(h^t, M^t, M^t)}
\end{equation}
$\widetilde{\mathrm{h_t}}$ will be passed to the subsequent modules 
in the Transformer layer. 

Similar approaches have been discussed in previous works. Simply increasing the context space does not help but introduces a lot of noise. Instead of incorporating multiple sentences to the context attention, we compress contextualized information into multiple memorized vectors and map such vectors back to the sentence state at the next step. We find that both the BLEU score and the information gained from the context attention space do not increase when the memory length increases from 64 to 128. Therefore, a large context space in $\mathrm{M}$ seems redundant for the model to learn, and we find $\mathrm{d_M}=16$ for the most effectiveness and efficiency.

\begin{table*}
\setlength{\tabcolsep}{4.5pt}
\centering
\begin{tabular}{lcccccc}
\hline
\textbf{Model} & \multicolumn{2}{c}{\textbf{TED}} & \multicolumn{2}{c}{\textbf{News}} & \multicolumn{2}{c}{\textbf{Europarl}} \\
\hline
 & s-BLEU  & d-BLEU & s-BLEU & d-BLEU & s-BLEU  & d-BLEU \\
\hline
\citet{NIPS2017_3f5ee243} & 23.10  & - & 22.40 & - & 29.40  & -  \\
\citet{miculicich-etal-2018-document} & 24.58 & - & 25.03 & - & 28.60 & -  \\
\citet{maruf-etal-2019-selective} & 24.42 & - & 24.84 & - & 29.75  & - \\
\citet{ma-etal-2020-simple} & 24.87 & - & 23.55 & - & 30.09 & - \\
\citet{ijcai2020-0551} & 25.10  & - & 24.91 & - & 30.40 & - \\
\citet{bao-etal-2021-g} & 
25.12 (+0.30) & 27.17 & 
25.52 (+0.33) & 27.11 & 
\textbf{32.39} (+1.02) & \textbf{34.08} \\

\hline
Sentence Baseline &  24.73 & - & 25.18 & - & 30.13 & - \\

Finetune on Sentence & 
\textbf{25.62} (\textbf{+0.89}) & \textbf{29.47} & 
\textbf{25.73} (\textbf{+0.55}) & \textbf{27.78}  & 
31.41 (\textbf{+1.28}) & 33.50 \\


\hline
\end{tabular}
\caption{Experiments results of BLEU scores on three datasets. The improvement from the Transformer baseline for previous models are also reported as in "()". It indicates the score improved from sentence-level translation provided by their implementations. Results are averaged from two runs.}
\label{table-experiments}
\end{table*}

\subsection{Document Neural Machine Translation}
In the task of document-level machine translation, the source and target documents are represented as sequences of sentences $\mathrm{X=\{x_t | 1\leq t \leq n\}}$, and $\mathrm{Y=\{y_t | 1\leq t \leq n\}}$ respectively, where $\mathrm{t}$ refers to the sentence index.  
Given a vanilla Transformer and its parameters $\mathrm{\theta}$, the objective is to maximize the target document probability conditioned on the source document.
\begin{equation*}
\mathrm{\argmax_{\theta} P(Y|X,\theta)}
\end{equation*}
Our approach recurrently translates an ordered document sentence by sentence, and the objective is:
\begin{equation*}
\mathrm{
    \argmax_{\widetilde{\theta}}
    \prod_{t} P(Y_{t}|X_{\leqslant{t}}, Y_{<t} ,\widetilde{\theta})
    }
\end{equation*}
where $\mathrm{\widetilde{\theta}}$ refers to Transformer parameters including Memory, Update Attention and Output Attention.

\begin{table}
\centering
\begin{tabular}{lcc}
\hline
\textbf{Data} & \textbf{\# of Docs} & \textbf{\# of Sents/Doc} \\
\hline
TED & 1.7K/93/23 & 123/98/105 \\
News & 6.1K/71/155 & 40/25/20 \\
Europarl & 118K/240/360 & 14/15/14 \\
\hline
\end{tabular}
\caption{Dataset Statistics for Train/Valid/Test}
\label{table-data}
\end{table}

As suggested by \citet{Beltagy2020Longformer, bao-etal-2021-g}, context would be better applied in higher layers and keep only local information in lower layers. Therefore we only apply the memory unit $\mathrm{M}$ in the top layer of encoder and decoder, and in lower layers, we keep using the original Transformer structure. Analysis regarding the location of memory is discussed in Section~\ref{memory}. 

\paragraph{Training} During training, our model takes an input of $\mathrm{x_t}$ and $\mathrm{y_t}$, which refer to sequences of tokens of the $\mathrm{t^{th}}$ sentence in source and target documents. Memory unit $\mathrm{M}$ is initialized trainable parameters before the first input of each document, and it will be updated after each input sentence pair, which are batched as the sentence order in the document. 
For computational convenience, the gradients are only back-propagated to the current and most recent sentences in each update step, and we stop the gradient for $\mathrm{M}$ before it is passed to the next step.

\paragraph{Inference}
In the decoding phase, our model translates the source document sentence by sentence. In the generation of each sentence, tokens are decoded in an auto-regressive order until the stop sign or exceeds the max length. The memory $\mathrm{M}$ will not be updated until the complete sentence is generated since the update of $\mathrm{M}$ depends on all tokens in the current sentence. If $\mathrm{M}$ is updated after each token generation, the attention space in the output attention does not represent the complete contextualized information of the expected sentence. 
The computational complexity keeps in sentence-level
since we only feed one sentence every time, and there is no cache vector besides $\mathrm{M}$.

\section{Experiment}

\subsection{Datasets}
We experiment across three widely used datasets for English$\to$German document translation. 

\setlength{\parindent}{2ex} \textbf{TED} Training data for TED comes from IWSLT'17. We use tst2016-2017 as test set and a held-out set from training as valid. 

\setlength{\parindent}{2ex} \textbf{News} The corpus comes from News Commentary v11. We use tst2016-2017 as test set and a held-out set from training as valid. 

\setlength{\parindent}{2ex} \textbf{Europarl} Train, valid and test sets are extracted from the corpus Europarl v7, as mentioned in \cite{maruf-etal-2019-selective}.

Detailed statistics for the datasets is in Table~\ref{table-data}. 
Moses \cite{koehn-etal-2007-moses} is used for data processing and 
BPE \cite{sennrich-etal-2016-neural} is used with vocab-size of 30K for all datasets.

\subsection{Settings}
We adopt Transformer model with the transformer-base configurations as the baseline, which has six layers with a hidden size of 512 and an intermediate size of 2048.
Token embedding is shared for source and target languages, and token indexes are encoded with a learnable embedding matrix. 
We first train a baseline model with vanilla Transformer architecture for sentence-level translation and finetune our model based on the sentence-level baseline.
We use the AdamW optimizer with an initial learning rate of $5\times10^{-4}$ and warm-up steps of 4000 for training sentence-level baseline. 
The drop-out rate is set to 0.3 for TED and News and 0.1 for the Europarl.
As for finetuning after sentence-level Transformer, the learning rate is set as $3\times10^{-4}$ for newly initialized parameters and $6\times10^{-5}$ for pretrained parameters, and warm-up steps of 1000 are set for TED and 2000 for News and Europarl.
The drop-out rate is set to 0.1 for the Europarl and 0.2 for the TED and News during finetuning.
We also apply gradient accumulation, and detailed studies are discussed in the section \ref{convergence}.
Models are trained with a patience of 5 for both sentence-level and document-level.
We use the beam size of 5 during inference and compute the BLEU score in a max order of 4 after removing BPE-tokens.
s-BLEU and d-BLEU are used as evaluation metrics, where s-BLEU refers to the BLEU score for sentences, and d-BLEU is the score for documents.

\begin{figure}[t]
\includegraphics[scale=0.35]{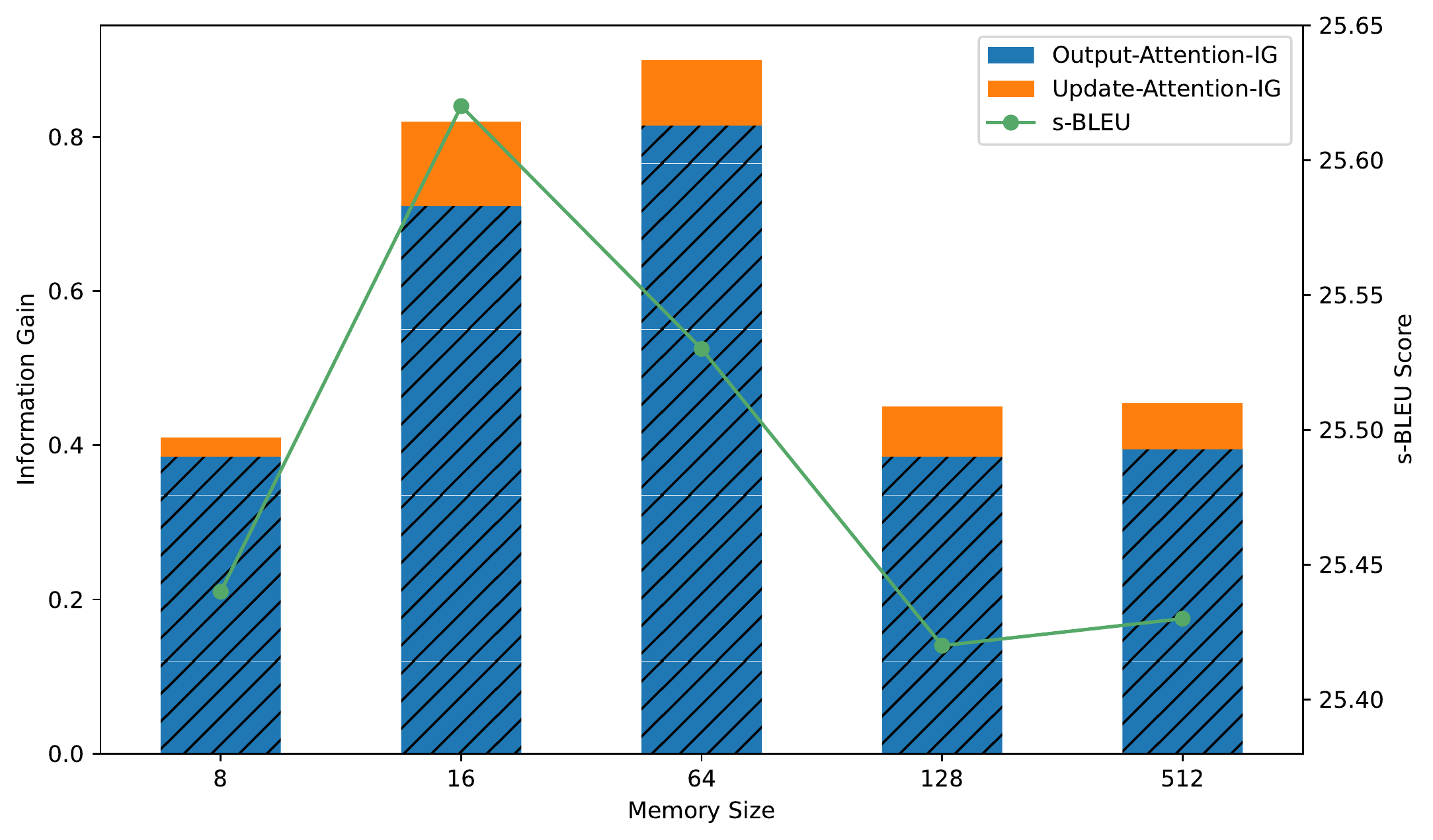}
\centering
\caption{Evaluation on TED with different memory sizes.}
\label{fig-memory-length}
\end{figure}

\subsection{Results}
Experiment results are shown in Table \ref{table-experiments}. 
Our method shows consistent improvements over three datasets from sentence-level Transformer.
We achieve the state-of-the-art results of s-BLEU of 25.62 and d-BLEU of 29.47 on TED and s-BLEU of 25.73 and d-BLEU of 27.78 on News . 
Though our results do not outperform G-Transformer \cite{bao-etal-2021-g} on Europarl, we think the difference mostly comes from the gap between sentence-level baselines.
Such difference may be caused by the implementation framework and computing resources, which they use Fairseq library and multiple GPUs, while we adopt the code from HuggingFace and only a single 1080-Ti GPU is used for our training.
We further report the score of works gained from their reported baseline, and our model makes the greatest improvement on all three datasets.
Overall, the results could demonstrate the advantages of our method in the general understanding of the document machine translation.


\section{Analysis}
In this section, we discuss our model from memory usage, long-range modeling, context effect, and computational complexity, respectively. 
Experiments are conducted with the model finetuned on the sentence baseline and evaluated on the TED, since TED has the most average sentence number per document, which is more likely to reflect the performance of our model for long documents.

\begin{table}
\centering
\begin{tabular}{lccc}
\hline
\textbf{ Side} & \textbf{Index}  & \textbf{s-BLEU} & \textbf{d-BLEU} \\
\hline
Source+Target & 0-1 & {25.31} & {29.13} \\
Source+Target & 2-3 & {25.30} & {29.23} \\
Source+Target & 4-5 & {25.43} & {29.22} \\
\textbf{Source+Target} & \textbf{5} & \textbf{25.62} & \textbf{29.47} \\
Source Only & 5 & 25.42 & 29.33 \\
Target Only & 5 & 25.43 & 29.25 \\
\hline
\end{tabular}
\caption{Evaluation on TED with memory on different sides and layers. We adopt the 6-layer Transformer model finetued on the sentence-level baseline, and 0 refers to the first layer, and 5 refers to the last layer.}
\label{table-mem-side-pos}
\end{table}

\subsection{Discussion of Memory}\label{memory}

\paragraph{Memory Size}
Memory size is evaluated through information gain (IG) between the random initialized memory and well trained memory. 
It is calculated from attention maps in Update Attention and Output Attention.
IG from Update Attention indicates the difference of selected information in the memory, and IG from Output Attention refers to how much contextualized knowledge in memory is mapped to the next sentence state. 
Figure \ref{fig-memory-length} shows IG keeps increasing as memory size increases from 8 to 64, but it dramatically drops at the size of 128 and 512.
While increasing the memory size can fit more contextual information, an excessively large memory space is likely to introduce redundant noise.
Therefore, it indicates that contextualized knowledge should be better distributed into a relatively dense space.
With the corresponding s-BLEU score, we set memory size to 16 in all other experiments for the most effectiveness.

\paragraph{Memory Side}
To analyze the effect of the memory on source and target documents, we set the memory on encoder, decoder and both sides respectively. We find that it is not only necessary to have the memory to convey the dependency between sentences on the source side but also in the decoding process for the target document. 
As shown in Table \ref{table-mem-side-pos}, applying the memory on either side can outperform the baseline but the model achieves better scores when incorporating the memory on both sides. It indicates the necessity of contextualized information for both source and target documents.

\paragraph{Memory Position}
Previous work \cite{bao-etal-2021-g, Beltagy2020Longformer} has shown that Transformer lower layers are more likely to have local information while the context is better incorporated into higher layers. We set the memory in lower, intermediate, and higher layers respectively. The results as shown in Table \ref{table-mem-side-pos} are consistent with the claim. Applying memory in higher layers outperforms the others, and it is even better to have it on only the top layer, which satisfies that the model is more likely to focus on the locality on lower layers and fuse the contextualized information on the top.

\subsection{Discussion of Long Dependency}
\begin{figure}[t]
\includegraphics[scale=0.35]{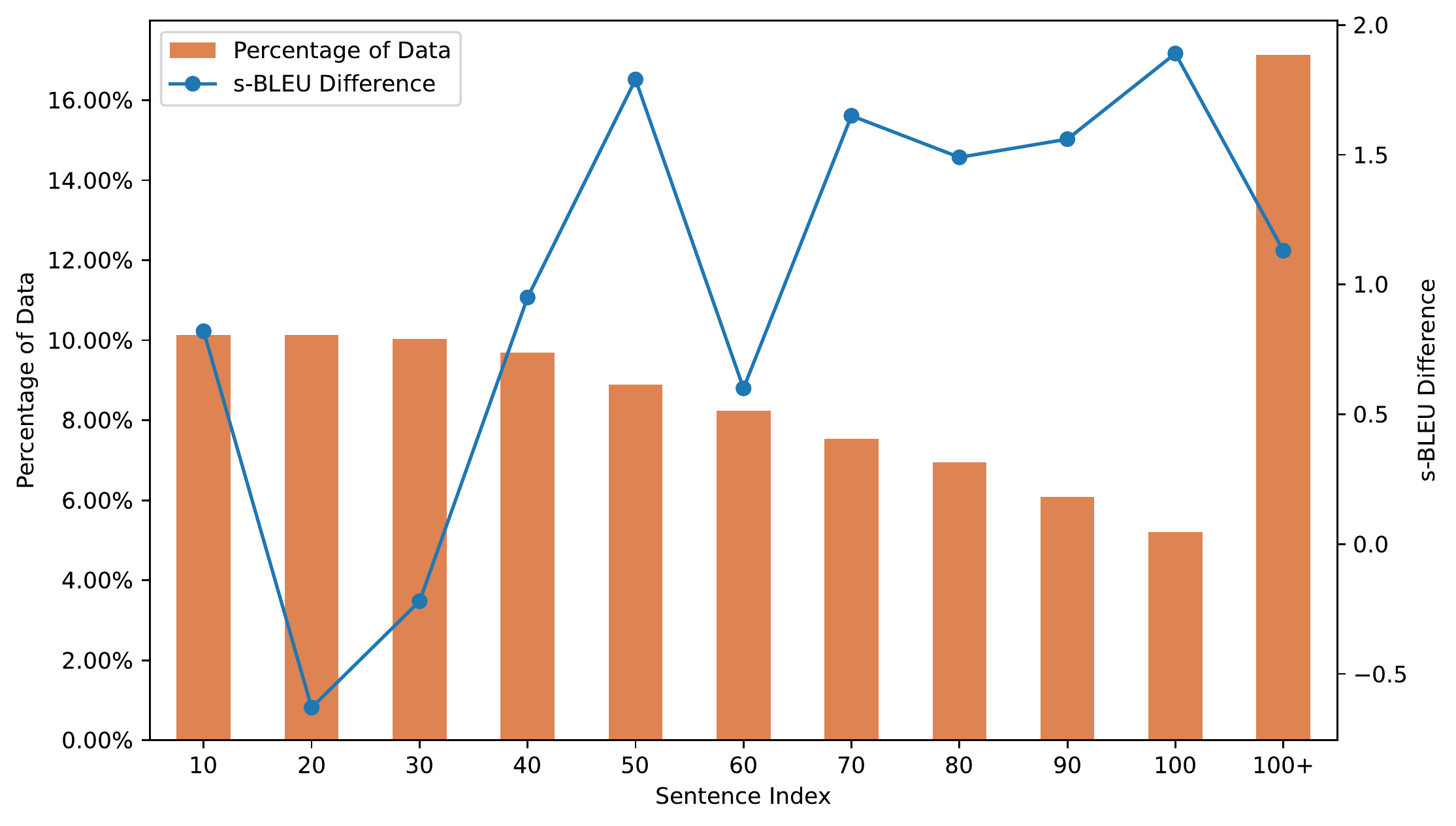}
\centering
\caption{TED datset separated by sentences from different indexes in documents, evaluated with Sentence-Transformer and Context-Aware Model.}
\label{fig-context-length}
\end{figure}

\paragraph{Metric Breakdown}
To find out on what kind of sentences our model outperforms the sentence-level Transformer, 
we evaluate the TED dataset with respect to the sentence index in the document. 
Sentences are ordered fed into the model.
We compute and average the s-BLEU for sentences at each sentence index in the document.
We further average the scores for every ten index range. 
As in Figure \ref{fig-context-length}, the x-axis refers to the index range of sentences (e.g., 20 refers to sentences with indexes from 10 to 20), and the y-axis indicates the s-BLEU difference between our model and sentence Transformer. 
Our model has consistently greater performance, especially for sentences in later part of documents,
indicating  our model has the superiority than the sentence-level Transformer on longer document translation and long-range modeling.

\begin{figure}[t]
\includegraphics[scale=0.35]{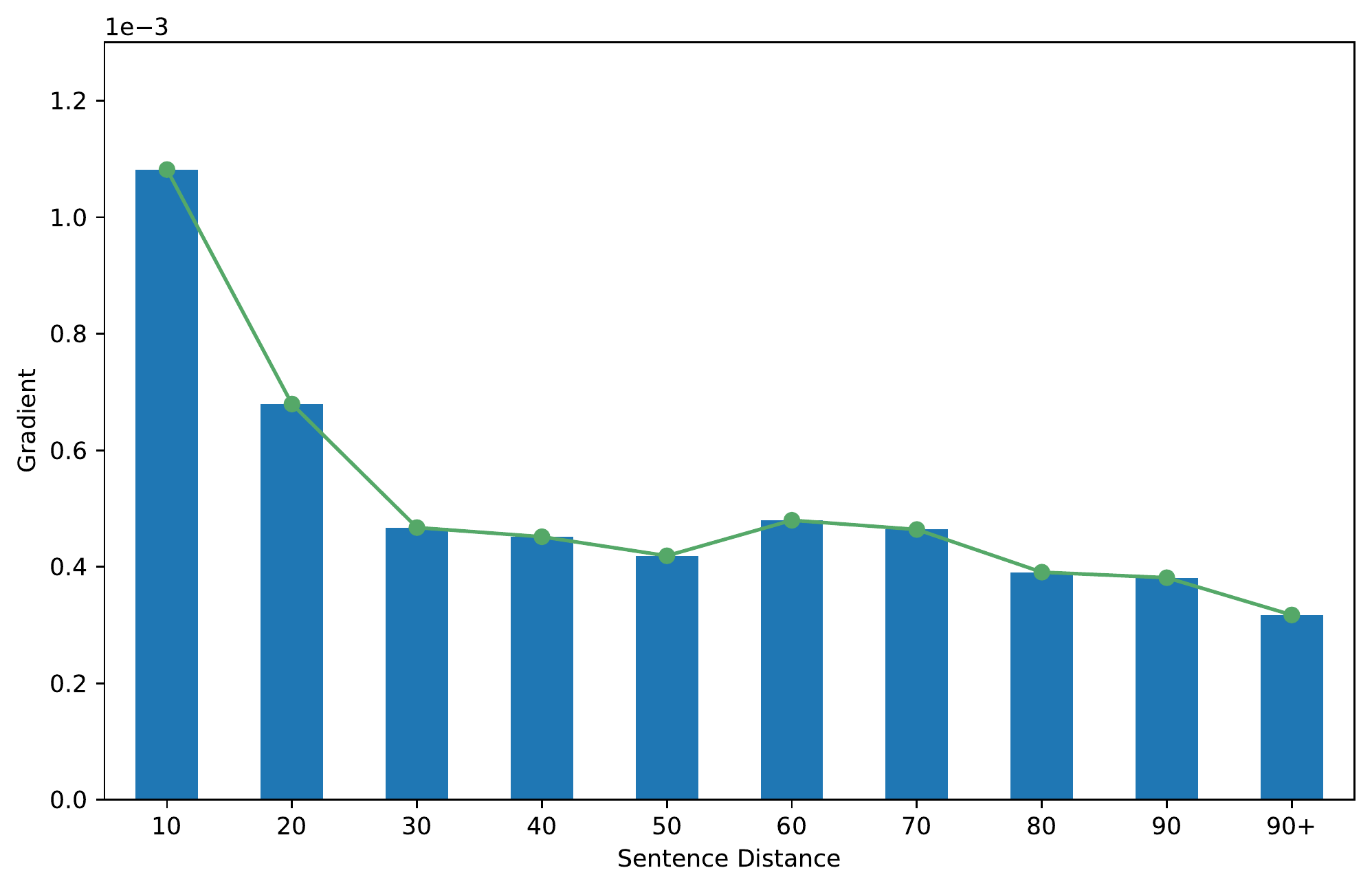}
\centering
\caption{TED dataset evaluated by gradients computed from different sentences  ranges. x-axis refers to the difference between the sentence indexes for gradient calculation and loss computation.}
\label{fig-gradient}
\end{figure}

\begin{figure*}[t]
\includegraphics[scale=0.21]{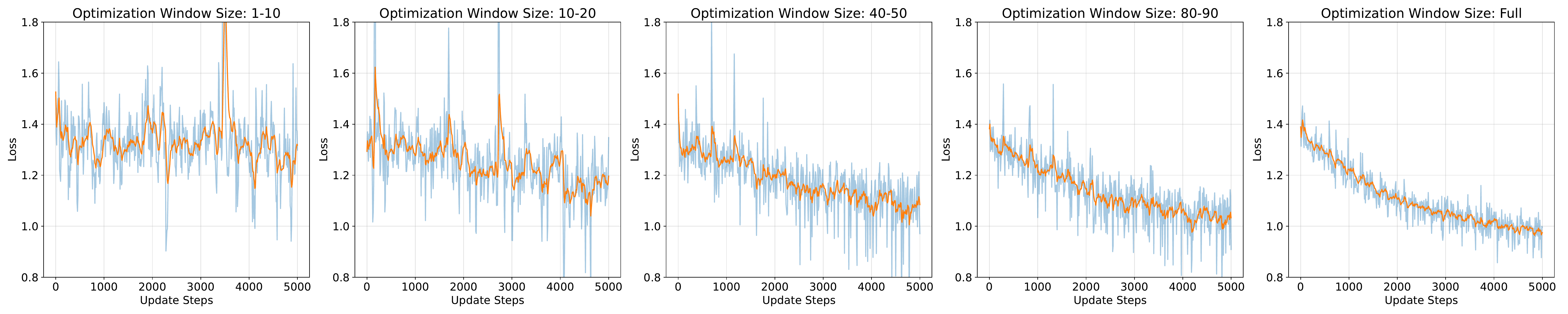}
\centering
\caption{Training Loss on TED Dataset, with different optimization window sizes}
\label{fig-opsize-loss}
\end{figure*}

\begin{figure}[t]
  \centering
\includegraphics[scale=0.26]{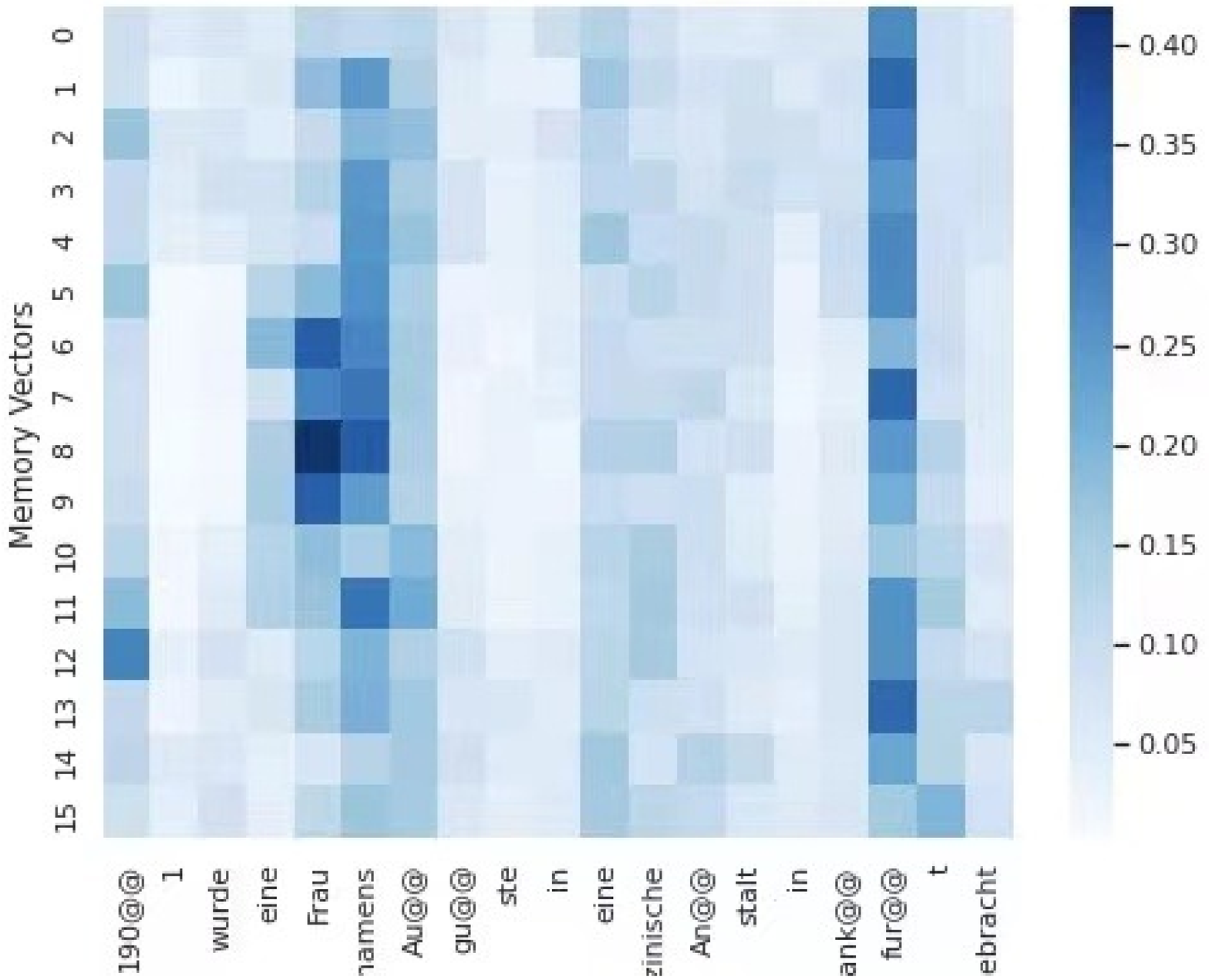}
\caption{Attention map from Update Attention, each token at sentence t is mapped to each memory vector.}
  \label{fig-update-att}
\end{figure}%
\begin{figure}[t]
  \centering
\includegraphics[scale=0.26]{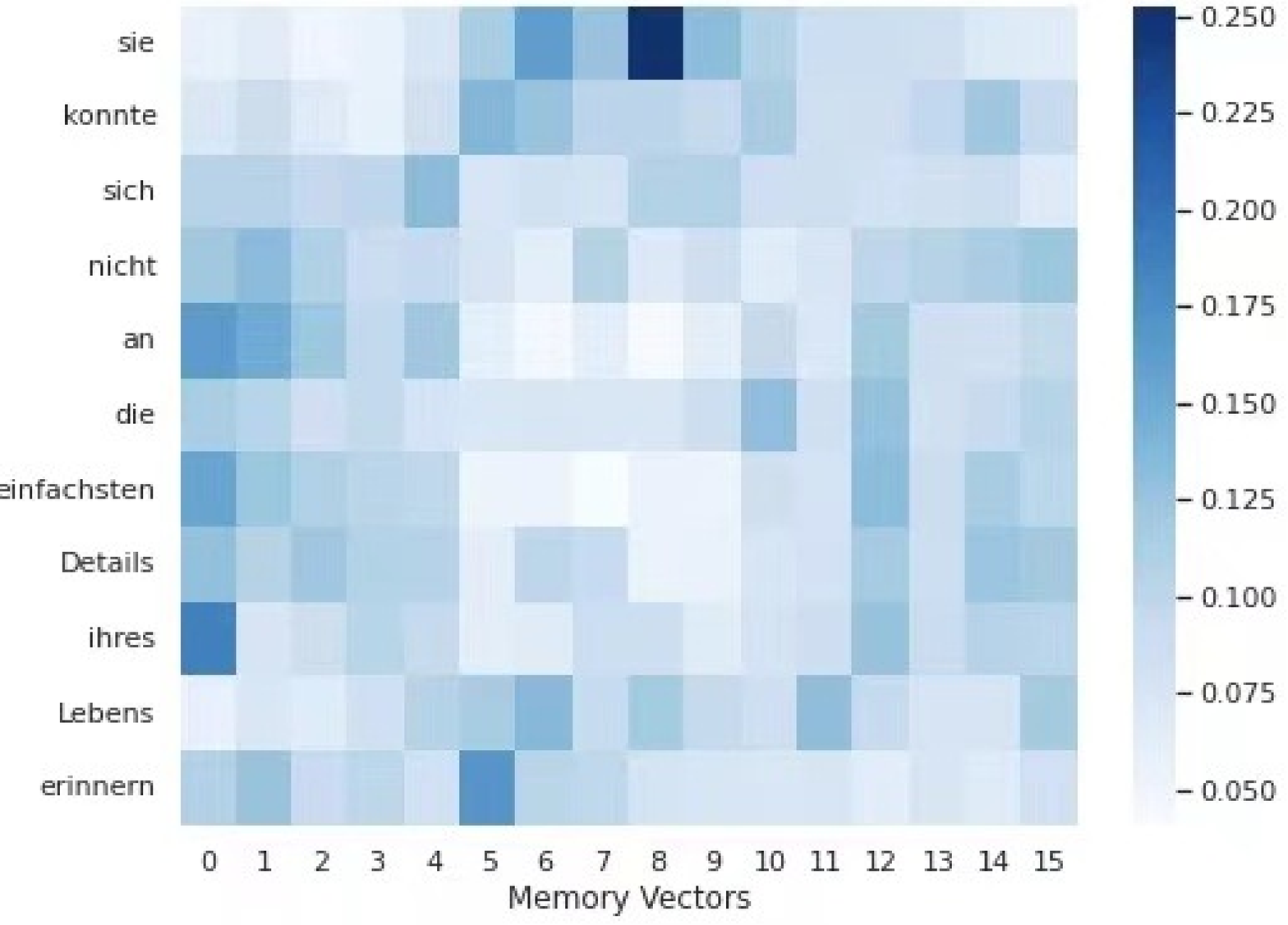}  
\caption{Attention map from Output Attention, memory vectors are mapped to each token in sentence t+1.}
  \label{fig-output-att}
\end{figure}

\paragraph{Long-Range Influence}
We also analyze the long-range dependency of our model through gradient attribution test introduced by \citet{ancona2018towards, pmlr-v70-sundararajan17a}. 
The gradient attribution test reflects the significance of the model input feature to its output prediction. 
We perform this test by calculating the gradients of our well-trained model on the test set of TED.
Since sentences are ordered when fed into the model, evaluating previous sentences' gradient attribution to the current sentence infers if the model supports the long-range dependency.
More formally, we define the gradient of the previous sentence $\mathrm{i}$ computed by the loss propagated from current sentence $\mathrm{j}$ as:
\begin{equation*}
\mathrm{G(Sent_i, Sent_j)}.
\end{equation*}

Specifically, the gradient of a certain token in previous sentences is retrieved from its corresponding embedding weight.
We conducted experiments for different sentence ranges $\mathrm{k}$ for the test with ten sentences intervals,
and the gradient for each range $\mathrm{k}$ is computed as:
\begin{equation*}
\mathrm{
Score(k) = Avg(\sum_{d=1}^{D}\sum_{s=1}^{S_d}\sum_{i=s+k}^{s+k+10}{G(Sent_s, Sent_i)}) 
}
\end{equation*}
where $D$ refers to number of documents, $\mathrm{S_d}$ refers to number of sentences in Document $\mathrm{d}$.
To prevent the gradient attribution accumulated by the same token within the evaluated range, only unique tokens within this range are considered.
As shown in Figure \ref{fig-gradient}, our model has gradients propagated to sentence tokens even by 90+ sentences from the computed loss, indicating our model does have the ability for long-range sequence modeling.

\subsection{Discussion of Context}
\paragraph{Convergence}\label{convergence}
Our model is trained concerning the sentence order in the document.
We find the model hard to converge during training as the loss oscillates within a wide range.
Because of the various distribution of consecutive sentences in documents, 
the directions of continuing optimization steps vary greatly, resulting in an unstable convergence curve.
To mitigate this issue, we use group optimization to update the model, considering the dependency among sentences. 
Specifically, a number from the optimization window is randomly sampled, and the gradients are accumulated.
The model will not be updated until the accumulated steps reach the sampled number.
We conduct experiments with different optimization window sizes for the update of 5000 steps, and the loss curves are shown in Figure \ref{fig-opsize-loss}, where full means the total number of sentences in the document. 
The result shows that the model converges faster and more stable with increasing optimization window size. Such improvement benefits from the grouped update steps concerning the difference of contextualized distribution among sentences.

\paragraph{Dependency Across Sentences}
We evaluate the attention maps from Update Attention and Output Attention to determine what contextualized information is passed in and out from memory. 
In Figure \ref{fig-update-att}, tokens from $\mathrm{t^{th}}$ sentence are mapped to each memory vector, and the $\mathrm{8^{th}}$ memory vector has a substantial attention weight on token "Frau".
Figure \ref{fig-output-att} shows memory vectors are mapped back to the following sentence and the token "sie" has a high probability on the $\mathrm{8^{th}}$ memory vector.
German words "Frau" and "sie" refer to "Mrs" and "she" in English. 
Hence, the memory mechanism has the ability to parse the word dependency between sentences at different steps.

\subsection{Discussion of Complexity}
\begin{figure}[t]
\includegraphics[scale=0.35]{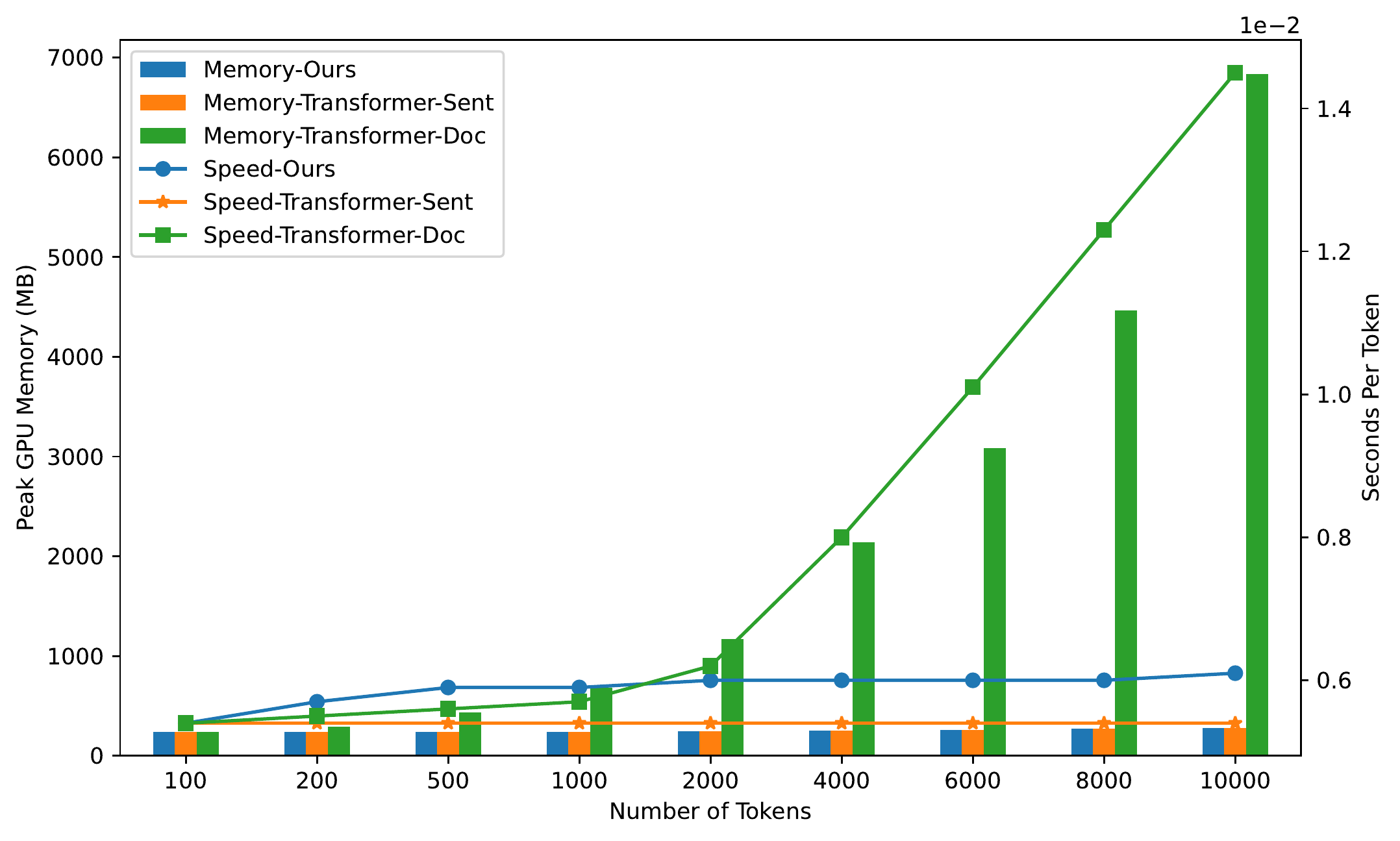}
\centering
\caption{Space and Time Complexity for different number of tokens during inference.}
\label{fig-complexity}
\end{figure}

We further analyze our model's space and time complexity during the decoding phase.
Since we only evaluate the decoding speed and memory efficiency in this case, we use dummy tokens to perform the inference.
We randomly generate a sequence of tokens as the source inputs and let the model decode the same number of tokens as the target.
We compare our model with both the sentence-level Transformer and document-level Transformer.
For the sentence-level Transformer, we split the sequence of tokens into chunks, and each chunk has a length of 100. 
The decoding complexity is evaluated over all chunks.
For the document-level Transformer, we use the entire sequence of tokens as the source input and evaluate the complexity of decoding the entire target sequence.
Similar to the sentence-level Transformer, our model is evaluated by the chunk by chunk decoding, and meanwhile, we keep the contextual memory updated.
As shown in Figure \ref{fig-complexity}, our model keeps the same space complexity as the sentence-level Transformer and takes a slightly more time cost because of the update of contextual memory.
However, the document-level Transformer has an increasing cost for both space and time complexity, especially when the target sequence has a length greater than 1,000 tokens. 
Overall, results have shown the decoding efficiency of our model, which keeps the computational complexity as low as the sentence-level Transformer, even in the case of over thousands of tokens.

\section{Conclusion}
This paper introduces a memory unit that recurrently maps information into and out of Transformer intermediate states and addresses the limitation about the context dependency and computational complexity in document-level machine translation. 
We have achieved the SOTA score on TED and News and a great improvement from the sentence-level baseline.
Our model demonstrates the effectiveness and efficiency of reduced memory space, context dependency for both source and target document, and long range influence across documents. 
The limitation of our work is the training cost since we accumulate the update steps and retain the graph for memory update at each step.
Our work does not conduct experiments for pretrained settings due to the time limitation. 
However, it should be easy to apply our method to any Transformer-based pretrained models, such as \citet{liu-etal-2020-multilingual-denoising}.
Also, this paper only experiments on document-level machine translation, and future works may apply this approach for other tasks that need long-range sequence modeling.

\section*{Acknowledgements}
We sincerely thank the reviewers from ACL Rolling Review for their helpful feedback, and the suggestions regarding figure plotting from Senior UI designer Xiangru Chen.
We also appreciate the support of the computing resources from the Center for Language and Speech Processing (CLSP) at Johns Hopkins University (JHU).

\bibliography{ref}
\bibliographystyle{acl_natbib}




\end{document}